\documentclass[10pt,twocolumn,letterpaper]{article}

\usepackage{cvpr}              % To produce the CAMERA-READY version
\definecolor{cvprblue}{rgb}{0.21,0.49,0.74}
\usepackage[pagebackref,breaklinks,colorlinks,allcolors=cvprblue]{hyperref}

\usepackage{amssymb}

\usepackage{multirow}
\usepackage{adjustbox} % Add this line
\usepackage{multirow} % Add this line at the beginning of your LaTe
\usepackage{makecell}
\usepackage{pifont}

\usepackage[utf8]{inputenc} % allow utf-8 input
\usepackage[T1]{fontenc}    % use 8-bit T1 fonts
\usepackage{url}
\usepackage{xurl}           % simple URL typesetting
\usepackage{hyperref}       % hyperlinks
\usepackage{booktabs}       % professional-quality tables
\usepackage{amsfonts}       % blackboard math symbols
\usepackage{nicefrac}       % compact symbols for 1/2, etc.
\usepackage{microtype}      % microtypography
\usepackage{xcolor}         % colors
\usepackage{graphicx} % for \includegraphics
\usepackage{subcaption} % for subfigure environment
\usepackage{wrapfig} % for wraptable environment
\usepackage{booktabs} % for \toprule, \midrule, \bottomrule
\usepackage{caption} % for \captionsetup

\usepackage{pifont}
\usepackage[dvipsnames]{xcolor}
\usepackage{colortbl}

\usepackage{graphicx}
\usepackage{booktabs}

\def\paperID{10175} % *** Enter the Paper ID here
\def\confName{CVPR}
\def\confYear{2026}

\title{From Inpainting to Layer Decomposition: \\ Repurposing Generative Inpainting Models for Image Layer Decomposition}

\author{
Jingxi Chen\textsuperscript{1, 2*}, Yixiao Zhang\textsuperscript{2}, Xiaoye Qian\textsuperscript{2}, Zongxia Li\textsuperscript{1},\\ Cornelia Fermuller\textsuperscript{1}, Caren Chen\textsuperscript{2}, Yiannis Aloimonos\textsuperscript{1}\\
\textsuperscript{1}University of Maryland \hspace{2cm}  \textsuperscript{2}Amazon\\ [2pt]
\url{https://inpaintinglayerdecomp.github.io/}
}

\begin{document}
% \maketitle

\twocolumn[{%
\renewcommand\twocolumn[1][]{#1}%
\maketitle
\begin{center}
    \captionsetup{type=figure}
    \includegraphics[width=\linewidth]{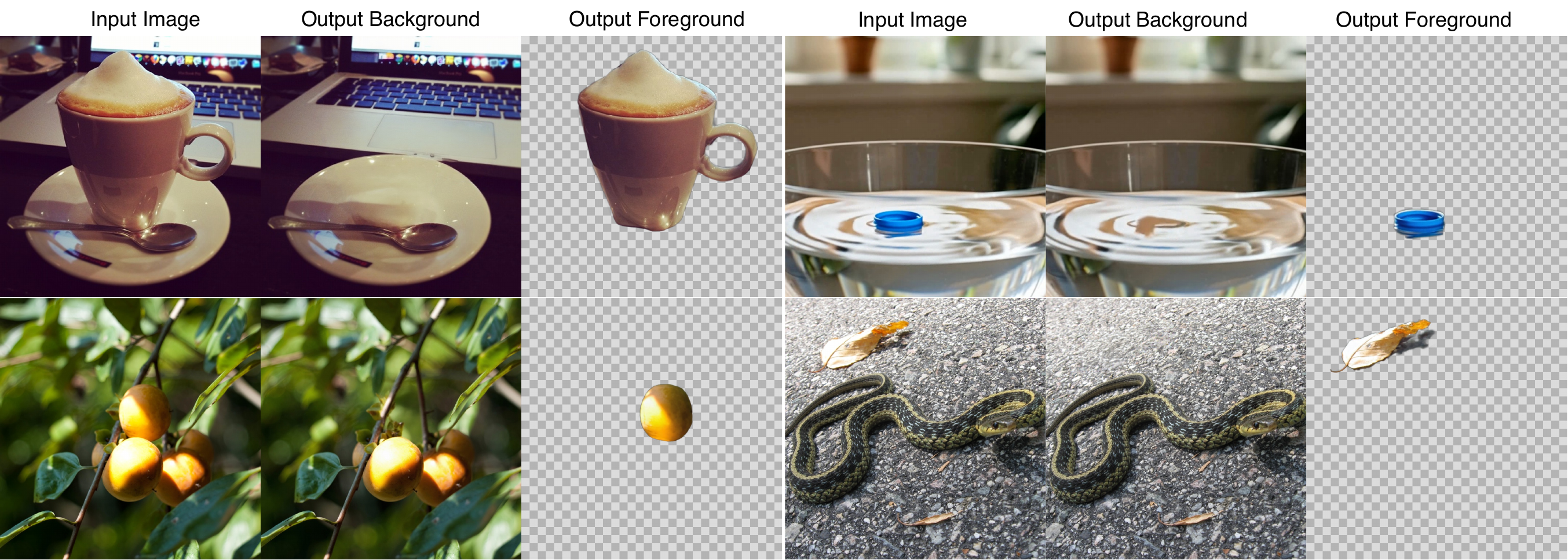} % Adjust the scale as needed
    \vspace{-20pt}
    \captionsetup{font=small}
    \caption{We propose a novel method for image layer decomposition, that is able to simultaneously extract foreground object with potential occlusion recovered, and remove the object from the background. The model is efficiently adapted from a pre-trained inpainting model. For each set of images in this figure, left is the original image, middle is the background and right is the foreground.}
    \label{fig:first}
    \vspace{-3pt}
\end{center}%
}]
\vspace{-10pt}

\begin{abstract}
% All images can be viewed as compositions of layered content -- foreground objects and background -- with potential occlusions in between. Layered representation of image allows independent editing of image elements, bringing much more flexibility for user content creation. Despite advancements in large generative models, the decomposition of a single image into layers remains challenging in terms of both methods and data. We observe the intrinsic similarity between the layer decomposition task and the in-painting/out-painting task, and propose that a layer decomposition model can be simply adapted from an diffusion-based in-painting model with lightweight finetuning. To further enhance the preservation of details in the latent space, we propose an innovative multi-modality context fusion module with linear attention complexity. We empirically verify that such a model can be learned purely on a synthetic dataset built with open-source resources. Our method achieves superior performance in both object removal and foreground occlusion recovery, enabling various possibilities in downstream editing and creation.

\begingroup
\renewcommand\thefootnote{}\footnote{*Work done during Jingxi Chen’s internship at Amazon}
\addtocounter{footnote}{-1}
\endgroup

Images can be viewed as layered compositions, foreground objects over background, with potential occlusions. This layered representation enables independent editing of elements, offering greater flexibility for content creation. Despite the progress in large generative models, decomposing a single image into layers remains challenging due to limited methods and data. We observe a strong connection between layer decomposition and in/outpainting tasks, and propose adapting a diffusion-based inpainting model for layer decomposition using lightweight finetuning. To further preserve detail in the latent space, we introduce a novel multi-modal context fusion module with linear attention complexity. Our model is trained purely on a synthetic dataset constructed from open-source assets and achieves superior performance in object removal and occlusion recovery, unlocking new possibilities in downstream editing and creative applications.
\end{abstract}    
\section{Introduction}
\label{sec:intro}

Recent advances in generative modeling, particularly diffusion models, have greatly improved image synthesis and manipulation. While most image diffusion models \cite{lugmayr2022repaint, shang2024resdiff, xia2024diffi2i} target single tasks like inpainting, super-resolution, or image translation, they excel at generating plausible content in under-constrained scenarios. However, real-world applications often require a combination of related tasks. One such task is image layer decomposition, simultaneously extracting the foreground and completing the background, which goes beyond inpainting and supports applications like creative re-composition, layered artwork generation, and component-based retrieval.

% \begin{figure}[!t]
%     \centering
%     \includegraphics[width=.999\linewidth]{sec/Figures/teaser.pdf}  

%     \caption{Given an input image and mask, pre-trained inpainting models typically focus only on background completion and may add unwanted content. Our adaptation enables simultaneous object removal, foreground extraction and outpainting, effectively repurposing the model for image layer decomposition. }

%     \label{fig:teaser}
%     \vspace{-8pt}
% \end{figure}

While single-task image diffusion models have been widely studied, diffusion-based image layer decomposition remains underexplored. A recent work~\cite{yang2025generative} addresses this by fully fine-tuning a closed-source, pre-trained text-to-image model on a large, curated dataset. Although effective, this approach demands substantial computational resources and data. In contrast, we observe that decomposed backgrounds involve filling masked regions, while foregrounds extend unmasked regions with an additional alpha channel. This conceptual similarity to inpainting and outpainting motivates our approach: to achieve layer decomposition through efficient adaptation of a pre-trained model, without large-scale training.

The central research question of this work is: \textit{What is the relationship between image layer decomposition and inpainting, and can existing inpainting models be effectively adapted to perform layer decomposition?}

This question drives our investigation into whether lightweight repurposing of inpainting models can offer a practical alternative to large-scale retraining for image layer decomposition. This direction is valuable to both researchers and practitioners, as training generative models from scratch is increasingly costly in terms of compute and data. In contrast, numerous pre-trained models already exist for standard tasks like inpainting. If these models can be effectively adapted for layer decomposition with minimal effort, it would significantly reduce development overhead and enable a plug-and-play extension that benefits from future advances in inpainting models.

In this work, we introduce the first approach to adapt a pre-trained image inpainting model for the novel task of image layer decomposition, as illustrated in Figure \ref{fig:first}. By examining the fundamental relationship between inpainting and layer decomposition, we observe that the latter extends inpainting by requiring not only background completion but also foreground extraction and outpainting. This insight motivates our method, \textit{Outpaint-and-Remove}, which efficiently adapts a pre-trained diffusion-based inpainting model to perform image layer decomposition in a parameter- and data-efficient manner.

For robust evaluation, we conduct experiments on the MULAN~\cite{tudosiu2024mulan} dataset, which is the only standard dataset in this domain to the best of our knowledge. We also conduct user study on a small set of real-word high quality images. Both qualitative and quantitative results show state-of-the-art performance on the novel task of image layer decomposition, as well as on standard object removal. Our ablation studies further validate the effectiveness of our design, demonstrating that augmenting a pre-trained inpainting model significantly improves its ability to perform layer decomposition.
Our main contributions are summarized as follows:
\begin{itemize}
    \item A novel perspective on layer decomposition, and a data- and parameter-efficient method that enables pre-trained image inpainting models to perform layer decomposition.
    \item Comprehensive experiments and ablations demonstrate that our adaptation-based approach effectively leverages publicly available models and datasets to achieve state-of-the-art performance on both image layer decomposition and the original inpainting task.
\end{itemize}

\section{Related Work}
\label{sec:related_work}
\subsection{Image Diffusion Models}
Diffusion models (or related flow-based models) \cite{rombach2022high, blattmann2023stable, peebles2023scalable, chen2025first, chen2025repurposing} have dominated image generation since the pioneering work DDPM~\cite{ho2020denoising}, outperforming VAE-based~\cite{kingma2013auto} and GAN-based~\cite{goodfellow2014generative} approaches in diversity, robustness, and scalability. While alternatives such as autoregressive models~\cite{tian2024visual} have been proposed, most state-of-the-art text-to-image models remain within the diffusion paradigm~\cite{batifol2025flux, rombach2022high}. Diffusion models have also been successfully adapted to a wide range of image tasks, including inpainting~\cite{lugmayr2022repaint, corneanu2024latentpaint, xie2023smartbrush}, super-resolution~\cite{gao2023implicit}, image translation~\cite{batifol2025flux, kwon2022diffusion}, and even discriminative tasks like segmentation and matting~\cite{amit2021segdiff, hu2024diffusion}.

\subsection{Image inpainting Models}
Early image inpainting methods typically rely on training convolutional neural networks or transformers from scratch on curated datasets~\cite{yu2018generative, yu2020region, suvorov2022resolution}. More recently, diffusion-based generative models have become the dominant approach for inpainting, driven by both industrial~\cite{sdxlinpainting, flux2024} and open-source~\cite{kandinsky2023} efforts. Compared to traditional methods, these models offer greater diversity through sampling and better controllability via prompt conditioning. 

% Our work builds on FLUX.1-fill-dev~\cite{flux}, one of the state-of-the-art diffusion-based inpainting models.

\subsection{Image Layer Decomposition}

Layer decomposition enables flexible element-wise image editing that is infeasible with single-layer images. However, high-quality layer-annotated datasets remain scarce in the open-source community. MULAN~\cite{tudosiu2024mulan} provides a large-scale dataset with RGBA layers, but these are generated using off-the-shelf detection, segmentation, or inpainting models, and their quality is not guaranteed. LayerDiffuse~\cite{zhang2024transparent} and Alfie~\cite{quattrini2024alfie} repurpose diffusion models to synthesize RGBA layers. In our work, we leverage MULAN and LayerDiffuse as foreground layer data sources. LAYERDECOMP~\cite{yang2025generative} is a recent method explicitly targeting the layer decomposition task. However, LAYERDECOMP requires full fine-tuning on a large scale high-quality dataset, making it impractical for general users with limited resources. In contrast, our method introduces lightweight LoRA-based fine-tuning strategy with multi-modal context inputs. This allows us to re-purpose existing pre-trained inpainting models for layer decomposition without large-scale training and democratize the capability to a broader range of users.
% Diffusion Transformer (DiT) on a close-sourced large-scale dataset, whereas we reveal the close relationship between inpainting and layer decomposition, and show that the latter can be achieved via lightweight fine-tuning on a pre-trained inpainting model, requiring significantly less data and computational cost.

\section{Proposed Approach}
\subsection{Problem Formulation} The core idea of our method is that image layer decomposition can be reformulated as a combination of inpainting and outpainting tasks. Rather than designing a model from scratch, we show that a single inpainting model can be efficiently fine-tuned to handle this task.

In this section, we discuss the conceptual connection between image inpainting and layer decomposition, and explain how we adapt the inpainting formulation to simultaneously recover missing regions (inpainting) and extend visible content beyond its original boundaries (outpainting), effectively enabling layered decomposition.

\textbf{Image inpainting:} The image inpainting problem can be formulated as follows: given an input image $I$ and a binary mask $M$, a diffusion-based inpainting model $f_{\theta1}$ takes the masked image $I \cdot M$ as input and predicts the missing content in the region indicated by $M$. Here, $\cdot$ denotes the element-wise masking operation that removes the target region in $I$ for the purpose of inpainting.

\textbf{Image Layer Decomposition:} For the Image Layer Decomposition, given an input image $I$ and a binary mask $M$,  a diffusion-based model $f_{\theta2}$ takes the image $I$ and $M$ as input and outputs both the extracted foreground indicated by $M$, which may include occluded regions, and a completed background with the foreground object removed. The key difference from image inpainting lies not only in the additional foreground output but also in the treatment of the background. As shown in Figure~\ref{fig:qualitative_1}, an inpainting model simply fills the masked region with plausible content, while layer decomposition requires explicit object removal, restoring the background as if the object never existed.

Nevertheless, we observe a structural unification between inpainting and layer decomposition tasks. For the background layer, both approaches involve filling masked regions with plausible content. For the foreground layer, the goal is to recover accurate RGB and alpha channels within the masked object, while enforcing zero alpha outside, effectively an outpainting formulation with transparency constraints. \textit{We also provide an intuitive illustration of the similarity between outpainting and foreground generation in the supplementary materials.}
This dual formulation allows both foreground and background generation to be viewed as special cases of in-/out-painting. As modern diffusion models are generalists due to large-scale pre-training, we propose reusing a single inpainting backbone with minimal modifications, adding a lightweight foreground prediction head, to support image layer decomposition via parameter-efficient fine-tuning.

\subsection{Pipeline Overview}
\begin{figure}[!t]
    \centering
    \includegraphics[width=.999\linewidth]{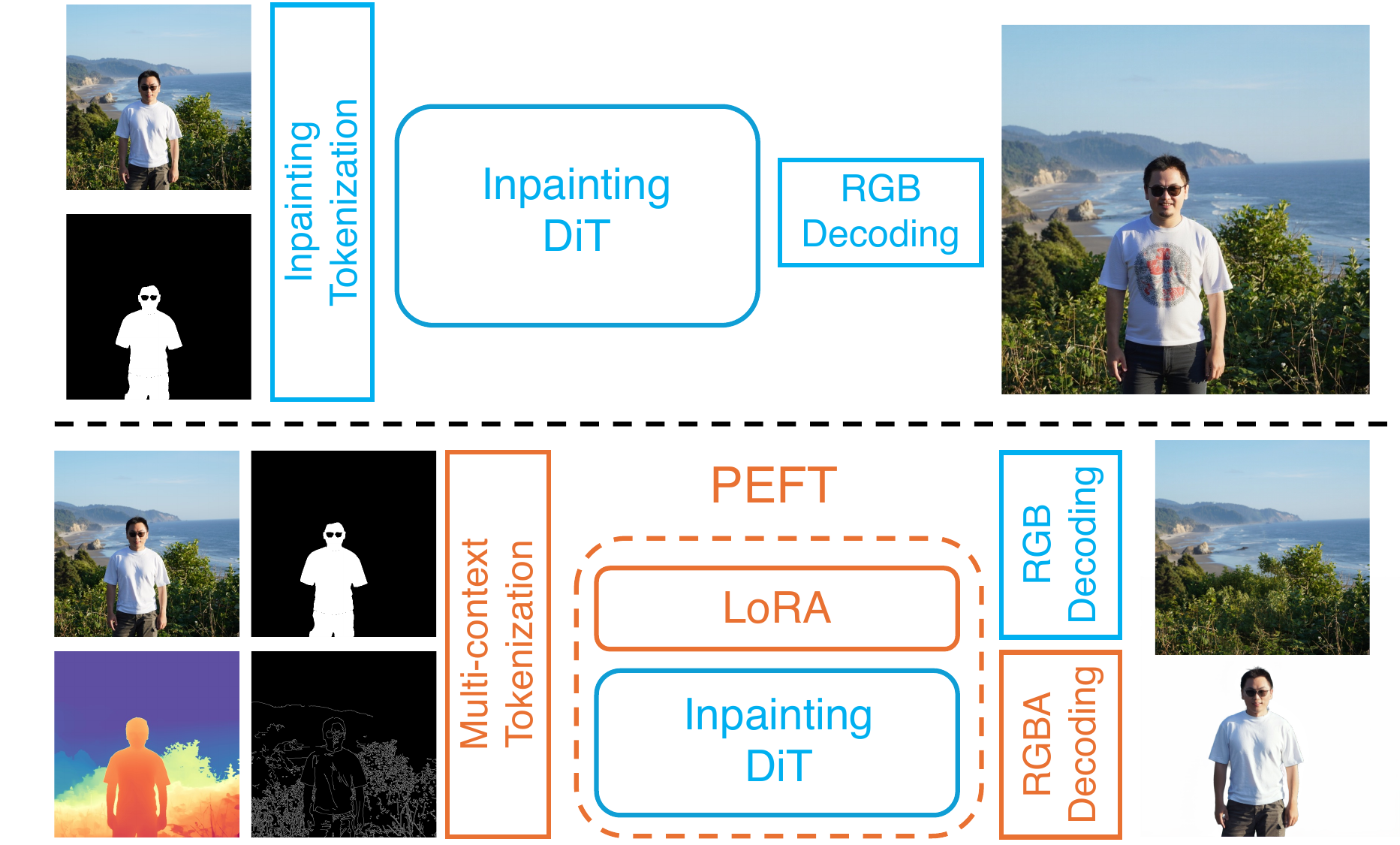}

    \vspace{-5pt}

    \caption{The top illustrates the original inpainting functionality of the pre-trained Inpainting DiT. The bottom shows our adapted pipeline for the image layer decomposition task. We introduce three key components to the pre-trained model for the adaptation: 1) Multi-Modal context Tokenization, 2) Parameter-Efficient Fine-Tuning (PEFT), 3) RGBA Decoding. Original components from the pre-trained model are highlighted in light blue, while our added or modified components are shown in orange. }
    \label{fig:pipeline}
    \vspace{-15pt}
\end{figure}

\begin{figure*}[!t]
    \centering
    \includegraphics[width=.99\linewidth]{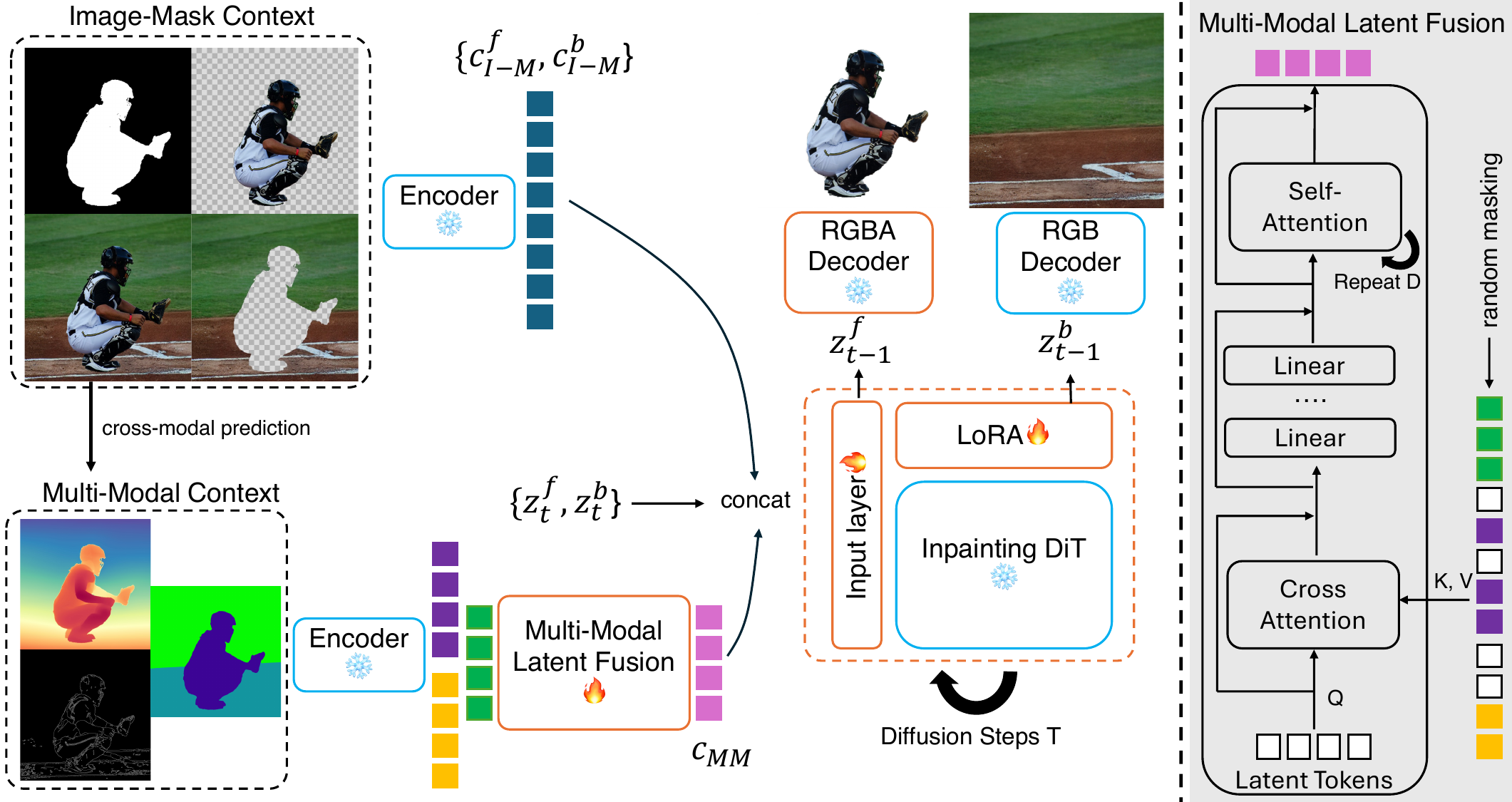}  

    \vspace{-6pt}
    \caption{Detailed diagram of the key components in our proposed adaptation method. Light blue boxes denote components from the original pre-trained inpainting DiT model, while orange boxes represent our modifications or additions. Our approach efficiently incorporates both Image-Mask Context and Multi-Modal Context tokens to guide generation. After adaptation, the model can simultaneously output an extracted and outpainted foreground along with a clean, object-removed background. }
    \label{fig:method}
    \vspace{-13pt}
\end{figure*}

We present an overview of our proposed Outpaint-and-Remove pipeline in Figure \ref{fig:pipeline}. In the Figure \ref{fig:method}, We detail our proposed method and key components. We also explain our unique data curation strategy built entirely from open-source datasets and tools in Figure \ref{fig:data_curatin}.

\subsection{Multi-Modal Context Tokenization}

Modern diffusion or flow models work on tokens in a latent space. To preserve as much fine details as possible in the latent space, we are motivated to exploit multi-modal context inputs as conditions for generation. Specifically, we produce the edge map, segmentation map and depth map from the given image.
% As a transformer-based architecture, the DiT takes context tokens as input and outputs predicted noise (or flow displacement) tokens at each diffusion step t. 
Given the image, mask, and multi-modal maps obtained, we first convert all these image-like inputs into tokens. With recent advances in VAE encoders from pre-trained DiT models, we find these encoders can effectively tokenize various image-like inputs, as verified in our supplementary material.

However, when we want to incorporate more modalities context tokens as additional inputs, challenge arises from the computational overhead introduced by the growing number of tokens. To address this, we propose summarizing the multi-modal tokens into a compact representation before feeding them into the DiT.
While token fusion is best handled by a transformer-like module, standard attention has quadratic complexity, $O(K^{2})$, where $K$ is the total number of tokens from all modalities. To reduce this cost, our proposed Multi-Modal Latent Fusion module, inspired by linear attention transformers~\cite{jaegle2021perceiver, mei2025power}, uses a small constant number $N << K$ of latent tokens as queries. This reduces the attention complexity to approximately $O(KN)$, making it linear with respect to the number of input tokens.

\subsection{Design of the Image-Mask Context}
\label{sec:image-mask}
The design of the image-mask context plays a critical role in our method. As shown in the ablation study (Section~\ref{sec:ablate-image-mask}), the choice of image-mask context directly reflects the nature of the task, a combination of foreground outpainting and background removal.

\begin{figure}[!t]
    \centering
    \includegraphics[width=.999\linewidth]{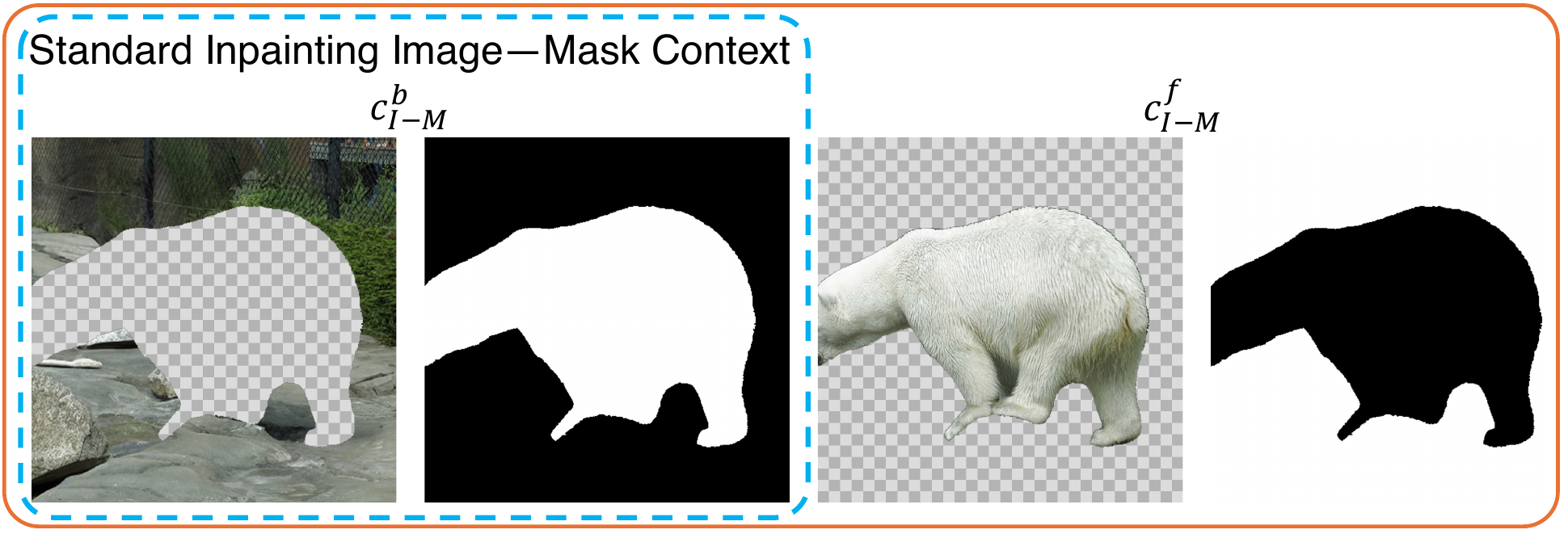}

    \vspace{-8pt}
    \caption{We illustrate the difference between the standard image-mask context used in diffusion-based inpainting models, shown in the light blue box as $c^{b}_{I-M}$, and our proposed image-mask context for image layer decomposition, shown in the orange box as $\{c^{f}_{I-M}, c^{b}_{I-M}\}$. }
    \label{fig:image-mask}
    \vspace{-18pt}
\end{figure}

Our context design is illustrated in Figure~\ref{fig:image-mask}. Unlike standard diffusion-based inpainting models that use only a background image-mask context $c^{b}_{I-M}$, we introduce an additional foreground image-mask context $c^{f}_{I-M}$. This foreground context serves as a control signal that helps the model balance preservation of existing information and generation of new content, enabling faithful extraction and outpainting of foreground objects.

After obtaining all context tokens from different sources, foreground context $c^{f}_{I-M}\in\mathbb{R}^{N \times d1}$, background context $c^{b}_{I-M} \in\mathbb{R}^{N \times d1}$, and fused multi-modal context $c_{MM}\in\mathbb{R}^{N \times d2}$, we concatenate them with the corresponding noisy input tokens for foreground $z^{f}_{t}\in\mathbb{R}^{N \times d3}$ and background $z^{b}_{t}\in\mathbb{R}^{N \times d3}$. The concatenation is performed along the channel dimension as follows: Foreground input:$concat(z^{f}_{t}, c^{f}_{I-M}, c_{MM}) \in  \mathbb{R}^{N \times (d1 + d2 + d3)}$, Background input: $concat(z^{b}_{t}, c^{b}_{I-M}, c_{MM}) \in  \mathbb{R}^{N \times (d1 + d2 + d3)}$. The final input to the DiT consists of two separate sequences of $N$ tokens, one for the foreground and one for the background, each with a token dimension of $d1 + d2 + d3$.

\subsection{Parameter Efficient Finetuning Protocol}
Inspired by prior work, we adopt a similar Parameter-Efficient Fine-Tuning (PEFT) protocol to adapt the frozen base inpainting DiT for our task. This allows the model to handle additional input channels and learn new capabilities efficiently with few training iterations. Specifically, we finetune the input projection layer and insert LoRA (Low-Rank Adaptation) layers into each attention and feed-forward layer of the DiT backbone.

Notably, as demonstrated in our ablation study (Section~\ref{sec:ablate-lora}), the LoRA rank plays a critical role in the detail preservation of foreground generation. It governs the trade-off between maintaining the generative power of the base inpainting model and learning the new capacity for foreground extraction, with the base model weights frozen.

\subsection{Decoding Foreground and Background}
The generated background is in RGB format, allowing us to directly reuse the pre-trained VAE encoder and decoder without modification. In contrast, the generated foreground is in RGBA format, which includes an additional alpha channel for transparency. To support this, we finetune a separate RGBA encoder/decoder, following the same principles as prior works~\cite{zhang2024transparent}.

\subsection{Training Data Curation from Public Sources}
\label{sec:data_curation}
In this section, we explain how we curate our synthetic training dataset using public sources and tools.
Our dataset includes key compositional elements commonly found in images, such as human, animals, objects, and indoor/outdoor scenes. 
% The inclusion of texts is particularly valuable for downstream artwork-related applications, such as poster design, textual editing and layer-aware re-targeting.

\begin{figure}[!t]
    \centering
    \includegraphics[width=.999\linewidth]{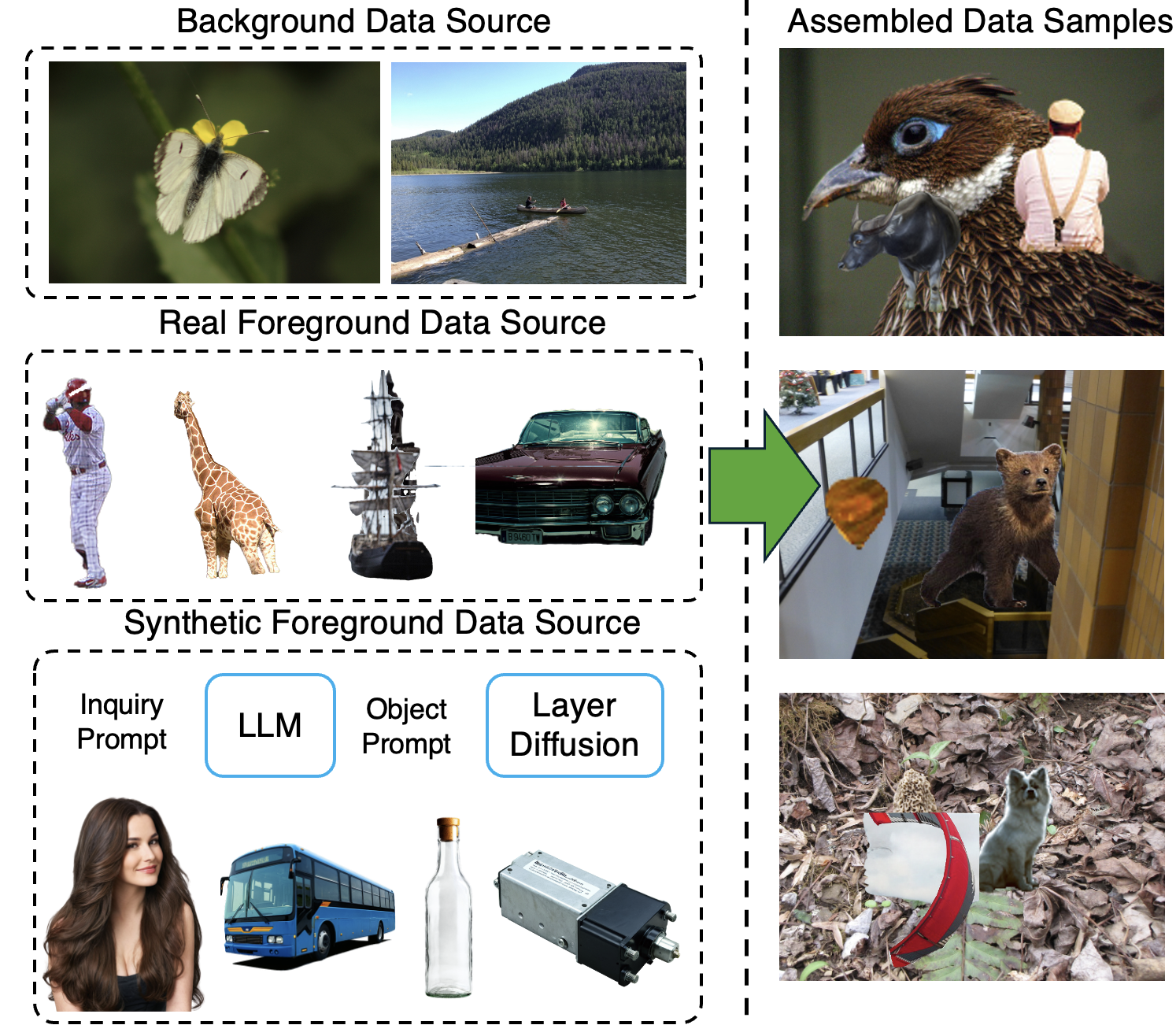}  

    \vspace{-8pt}
    \caption{Our training data consists three sources: backgrounds, real foregrounds, and synthetic foregrounds.}
    \vspace{-20pt}
    \label{fig:data_curatin}
    
\end{figure}

We use the MULAN~\cite{tudosiu2024mulan} dataset, a real-world layered image dataset, as one of the foundation for RGBA foreground layers. However, as shown in Figure~\ref{fig:data_curatin}, images in MULAN often suffer from imperfect foreground extraction due to occlusions and background entanglement. To address this, we explore a text-to-layer generative model LayerDiffuse~\cite{zhang2024transparent} to generate cleaner foregrounds. While these models produce complete foreground object shapes, they tend to introduce over-smoothed textures and lack fine details, leading to degraded training performance.

To harness the strengths of both sources of images, we propose a hybrid data strategy that combines two complementary foreground sources:1) Real foregrounds with rich detail but incomplete shapes, 2) Generated foregrounds with complete shapes but limited texture fidelity. By merging these two types, we construct a more balanced and effective training dataset.
% that significantly improves model performance, as confirmed in our ablation study.
\section{Experimental Results}

\subsection{Datasets and Implementation Details}
For the base inpainting model, we adopt FLUX.1-Fill-dev \cite{flux} as the pre-trained diffusion transformer (DiT)\footnote{We use it only for research purpose and abide by their Non-Commercial License.}. The default LoRA rank for training is 256. The input resolution is 1024x1024. The edge map is produced by Canny~\cite{canny2009computational}, the segmentation map is from SegFormer~\cite{xie2021segformer}, and the depth map is from Depth-Anything-V2~\cite{yang2024depth}. We trained the model using standard flow matching loss with batch size 8, learning rate $5e-5$ for 7200 iterations. Our curated training data is constructed entirely from public datasets and tools, ensuring both accessibility and reproducibility. 
Layered image datasets based on real images are extremely limited. To the best of our knowledge, MULAN~\cite{tudosiu2024mulan} is the only standard dataset in this domain. For foreground objects, we use real-world layered image from MULAN, as the primary source. We incorporate generated foreground layers from LayerDiffuse~\cite{zhang2024transparent}, guided by foreground prompts generated via ChatGPT-4o. We include examples of such prompts in the Supplementary Material. For background images, we sample from OpenImages~\cite{kuznetsova2020open}, which are then overlayed by rescaled foregrounds to simulate realistic composite scenes. Each training image contains 1–3 foreground objects with possible occlusions. We intentionally use imperfect masks for training so the model learns to infer the accurate object boundary even in the cases of inaccurate input masks. In total, we assemble 100,000 image–foreground–background triplets, constructed entirely from public sources.

\subsection{Evaluation Strategy}
We compare our method against several state-of-the-art (SOTA) object removal baselines, including SD-XL Inpainting~\cite{sdxlinpainting}, PowerPaint~\cite{zhuang2024task}, BrushNet~\cite{ju2024brushnet}, OmniEraser \cite{wei2025omnieraser}, GeoRemover \cite{zhu2025georemover}, Qwen-Image-Edit \cite{wu2025qwen} and our base pre-trained model, FLUX.1-Fill-dev \cite{batifol2025flux}. Quantitative evaluations are conducted on the 526 images from test sets of MULAN. Additionally, we perform qualitative comparisons on a diverse set of real-world images across various subjects and application domains to assess the quality of decomposed layers. 

For evaluation, we employ standard metrics widely used in image reconstruction and object removal tasks, including PSNR, SSIM \cite{wang2004image}, LPIPS \cite{zhang2018unreasonable} and FID \cite{heusel2017gans}. To further validate the practical effectiveness of our method, we also include results from a user study on the foreground generation quality, reflecting subjective human preferences and alignment with real-world use cases.

\subsection{Qualitative and Quantitative Results}

We report quantitative results on the MULAN test set to demonstrate the effectiveness of our method compared with baselines, as shown in Table~\ref{tab:metric}. The metrics are computed on background layers. Our adapted model achieves the best performance across all evaluation metrics, confirming the effectiveness of our approach. Furthermore, when compared to our base FLUX.1-Fill-dev model, our adaptation yields substantial improvements in background object removal, with a gain of 1.71 dB in PSNR and a reduction of 9.99 in FID.

\begin{figure*}[!t]
    \centering
    \includegraphics[width=.999\linewidth]{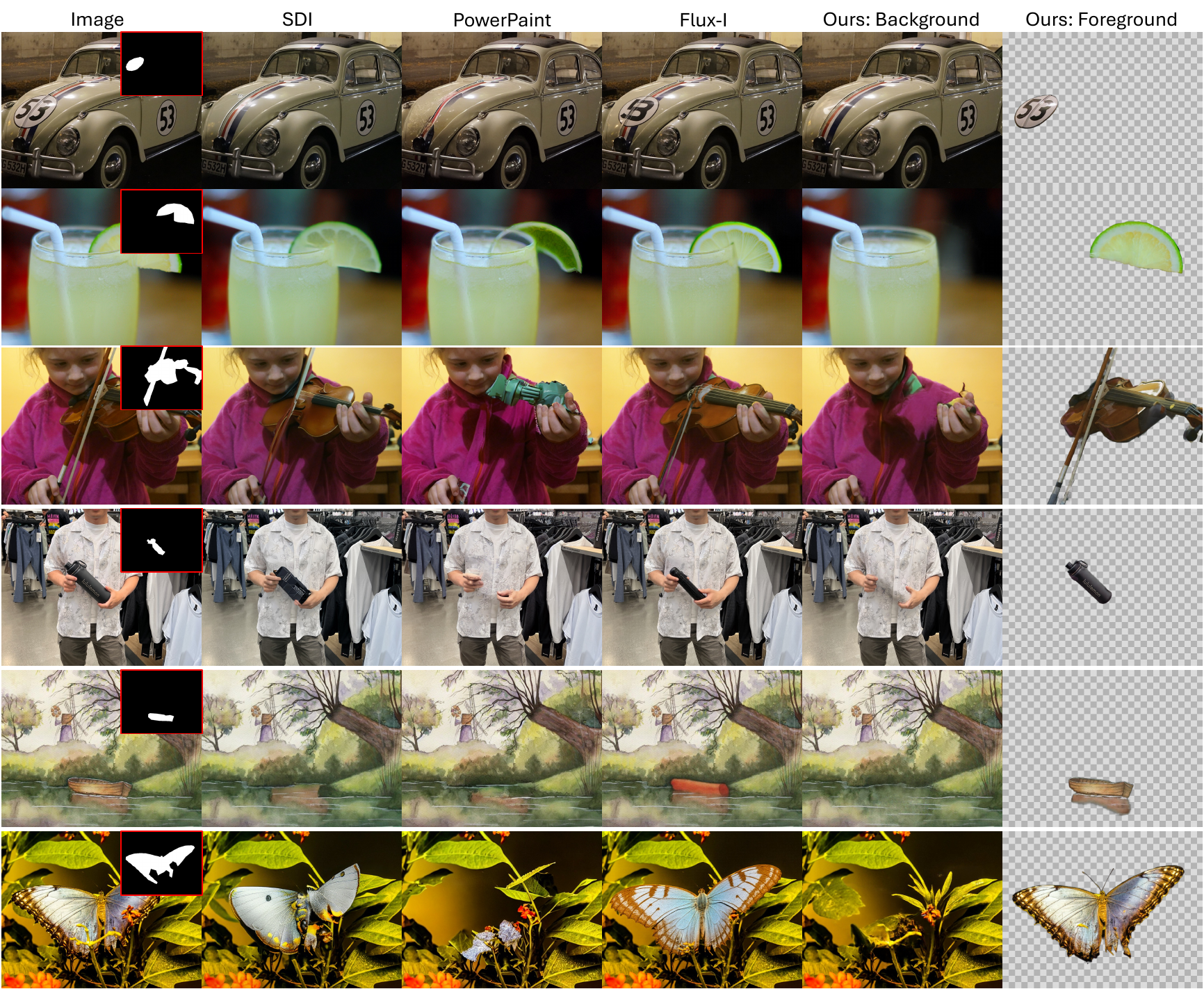}  
    \vspace{-20pt}
    \caption{We present examples comparing our method against baselines on our collected real-world image test set for the object removal task. These qualitative results highlight the visual differences in foreground removal accuracy, background reconstruction quality, and consistency across various challenging scenes. Please zoom in for the best viewing quality.}
    \label{fig:qualitative_1}
    \vspace{-8pt}
\end{figure*}

\begin{figure*}[!t]
    \centering
    \includegraphics[width=.999\linewidth]{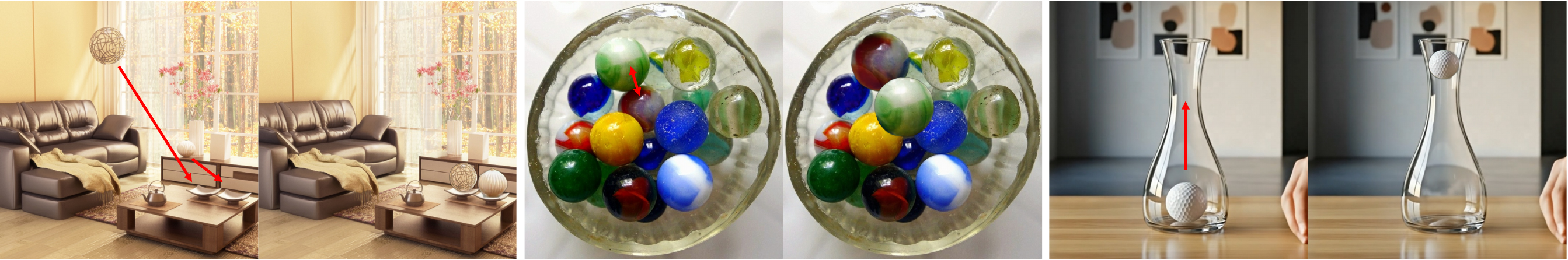}  
    \vspace{-20pt}
    \caption{Image editing examples. Layer decomposition enables manipulation of elements. Left: the original image, right: the edited image.}
    \label{fig:qualitative}
    \vspace{-14pt}
\end{figure*}

% However, in MULAN, the ground truth background layers are generated using learning-based methods, rather than captured from actual background images.
To better evaluate our model’s performance, and to fairly compare against baselines for the foreground layers on real-world images, we collect 40 diverse high-quality real-world images. Since there is no open-source methods dedicated to layer decomposition, we compare with two matting methods that produce RGBA foregrounds, MattingAnything~\cite{li2024matting} and DiffMatte~\cite{hu2024diffusion}. We also conducted a user study with 18 independent researchers to assess the fidelity of the extracted foreground layers. We present the quantitaive results in Table \ref{tab:user_study}. We also include some examples of qualitative comparison results in the Figure \ref{fig:qualitative_1}, image editing examples in Figure~\ref{fig:qualitative} and more in the supplementary materials.

% \begin{table}[t!]
%     \centering
%     \resizebox{1.0\linewidth}{!}{
%     \begin{tabular}{@{}ccccc@{}} % Four columns
%     \midrule
%     \midrule
%     & \makecell{\textbf{Method}} & Superior & Comparable & Inferior \\
%     \midrule
%      & Matting-Anything & 2.50\% & 2.50\% & 95.00\% \\
%      & DiffMatte & 25.00\% & 0.00\% & 72.5\% \\
%      & Ours & 70.00\% & 2.50\% & 27.50\% \\
%     \midrule
%     \midrule
%     \end{tabular}}

%     \captionsetup{font=small}
%     \caption{We performed a user study to compare the foreground generated with two matting baselines. The evaluation was carried out on our collected set of diverse real-world test images, allowing us to assess the practical effectiveness and visual quality of each method from the perspective of professional end users. For each image, we use majority voting to decide which model is superior, and if two models get the same top votes, they are regarded comparable for this image.}
%     \label{tab:user_study}

%     % \vspace{-6mm}
% \end{table}

\begin{table}[t!]
    \centering
    \resizebox{1.0\linewidth}{!}{
    \begin{tabular}{@{}cccccc@{}} % Four columns
    \midrule
    \midrule
    & \makecell{\textbf{Method}} & PSNR $\uparrow$  & SSIM $\uparrow$  & LPIPS $\downarrow$ & FID $\downarrow$  \\
    \midrule
     % & CNI & xxx& xxx & xxx \\
     & SDI & 20.92 & 0.84 & 0.17 & 69.93 \\
     & PowerPaint & 23.46 & 0.76 & 0.17 & 41.67 \\
     & BrushNet & 21.53 & 0.85 & 0.19 & 88.46 \\
    & OmniEraser & 21.45 & 0.72 & 0.31 & 55.80 \\
    & GeoRemover & 17.19 & 0.76 & 0.32 & 92.77 \\
    & Qwen-Image-Edit & 19.07 & 0.64 & 0.24 & 63.49 \\
    \midrule
     &  FLUX.1-Fill-dev & 25.59 & 0.92 & 0.09 & 35.96\\
     & Ours & \textbf{27.30} & \textbf{0.93} & \textbf{0.08} &\textbf{25.97} \\
    \midrule
    \midrule
    \end{tabular}}

    \vspace{-9pt}

    \captionsetup{font=small}
    \caption{Quantitative comparison of background removal.}
    \vspace{-10pt}
    
    \label{tab:metric}

    %\vspace{-6mm}
\end{table}

\begin{table}[t!]
    \centering
    \resizebox{1.0\linewidth}{!}{
    \begin{tabular}{@{}ccccccc@{}} % Four columns
    \midrule
    \midrule
    & \makecell{\textbf{Method}} & PSNR $\uparrow$  & SSIM $\uparrow$  & LPIPS $\downarrow$ & FID $\downarrow$ & User Preference Rate $\uparrow$ \\
    \midrule
     % & CNI & xxx& xxx & xxx \\
    & Matthing-Anything & 25.11 & 0.95 & 0.09 & 56.61 & 8.15\% \\
    & DiffMatte & \textbf{29.69} & \textbf{0.98} & \textbf{0.05} & \textbf{28.50} & 32.34\%  \\
    & Ours & 28.38 & 0.97 & 0.07 & 39.52 & \textbf{59.51\%} \\
    \midrule
    \midrule
    \end{tabular}}

    % \captionsetup{font=small}
    \vspace{-9pt}
    \caption{Quantitative comparison of foreground extraction. In addition to image quality metrics, we also conducted a user study comparing our foreground generation with two matting baselines on diverse real-world test images, evaluating practical effectiveness and visual quality from the perspective of professional end users.}
    \vspace{-15pt}
    \label{tab:user_study}
\end{table}

% \begin{table}[t!]
%     \centering
%     \resizebox{1.0\linewidth}{!}{
%     \begin{tabular}{@{}ccccc@{}} % Four columns
%     \midrule
%     \midrule
%     & \makecell{\textbf{Method}} & Matting-Anything & DiffMatte & Ours \\
%     \midrule
%      & User Preference Rate & 8.15\% & 32.34\% & 59.51\% \\
%     \midrule
%     \midrule
%     \end{tabular}}

%     \captionsetup{font=small}
%     \caption{We performed a user study to compare the foreground generated with two matting baselines. The evaluation was carried out on our collected set of diverse real-world test images, allowing us to assess the practical effectiveness and visual quality of each method from the perspective of professional end users.}
%     \label{tab:user_study}

%     % \vspace{-6mm}
% \end{table}

\subsection{Ablation Study}
\label{sec:ablation}
To evaluate the effectiveness of our design in adapting a pre-trained image inpainting diffusion model for the image layer decomposition task, we conduct a comprehensive ablation study as shown in Table \ref{tab:abla}. This analysis underscores the contribution of each design component, demonstrating their roles in balancing generative capability with faithful preservation of input details and contextual information.

\subsubsection{The Impact of Base Model and Training Data}
To investigate the relationship between image inpainting and layer decomposition, we compare two base model choices: a pre-trained inpainting model (FLUX.1-Fill-dev) and a general image-to-image model (FLUX.1-Kontext-dev). As shown in Table~\ref{tab:abla} :\textit{Kontext}, switching to the FLUX.1-Kontext-dev backbone leads to a performance drop, supporting our hypothesis that layer decomposition is best approached as an adaptation of the inpainting task. As expected, Table~\ref{tab:abla} also shows that incorporating synthetic foregrounds improves performance, validating one of the key advantages of our training data design.

\subsubsection{The Impact of Image-Mask Context}
\label{sec:ablate-image-mask}
\begin{figure}[!t]
    \centering
    \includegraphics[width=.999\linewidth]{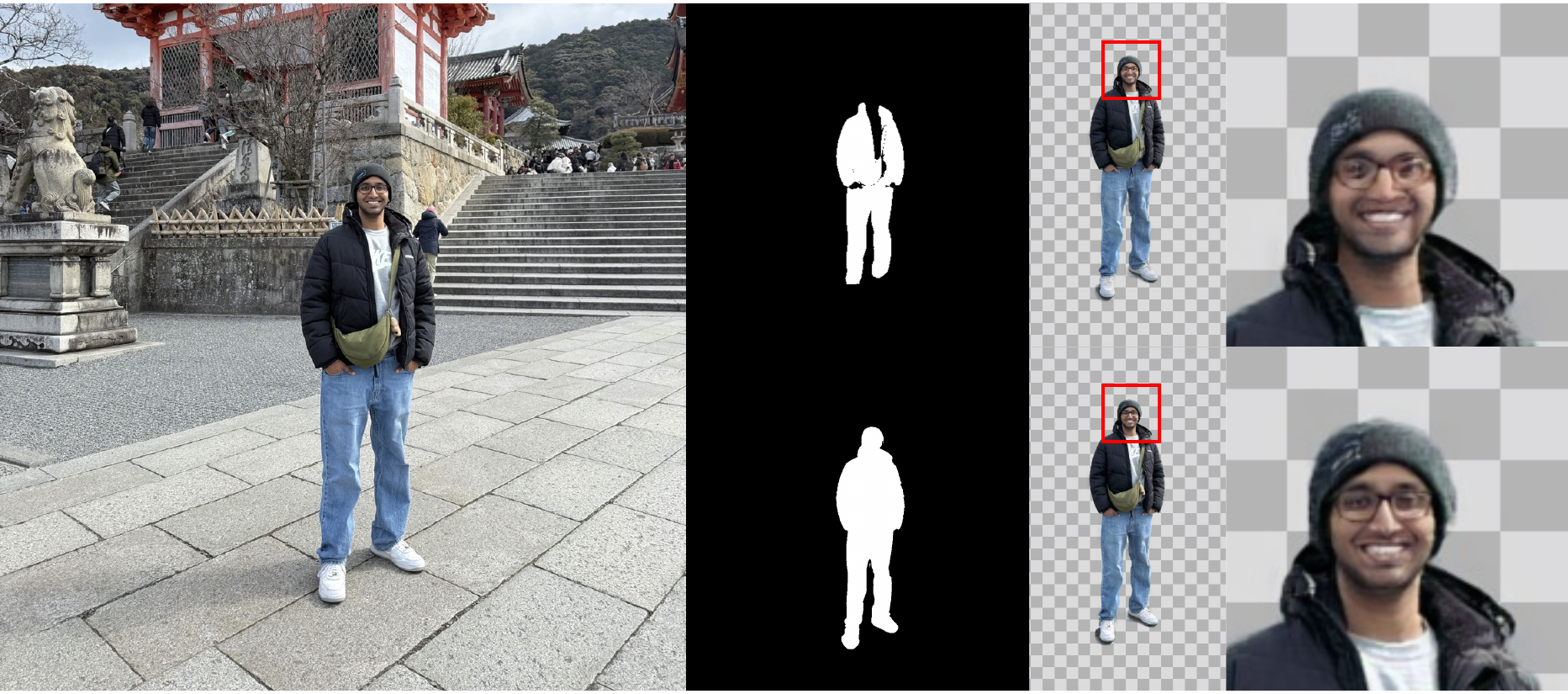}  
    \vspace{-20pt}
    \caption{We compare foreground generation results from the same input image using two different masks. The top row uses an incomplete mask that excludes the human head, causing the model to hallucinate or alter that region. The bottom row uses a nearly complete mask that includes the head, allowing the model to preserve it faithfully. This illustrates the importance of the input mask in balancing generation and content preservation.}
    \label{fig:abla_mask}
    \vspace{-9pt}
\end{figure}

As shown in Figure~\ref{fig:abla_mask}, the image-mask context plays a crucial role in our model by distinguishing between regions intended for generation and those meant for preservation. This allows the model to effectively control hallucinations arising from the inherent tension between completing missing content and faithfully retaining input details.

\begin{figure}[!t]
    \centering
    \includegraphics[width=.999\linewidth]{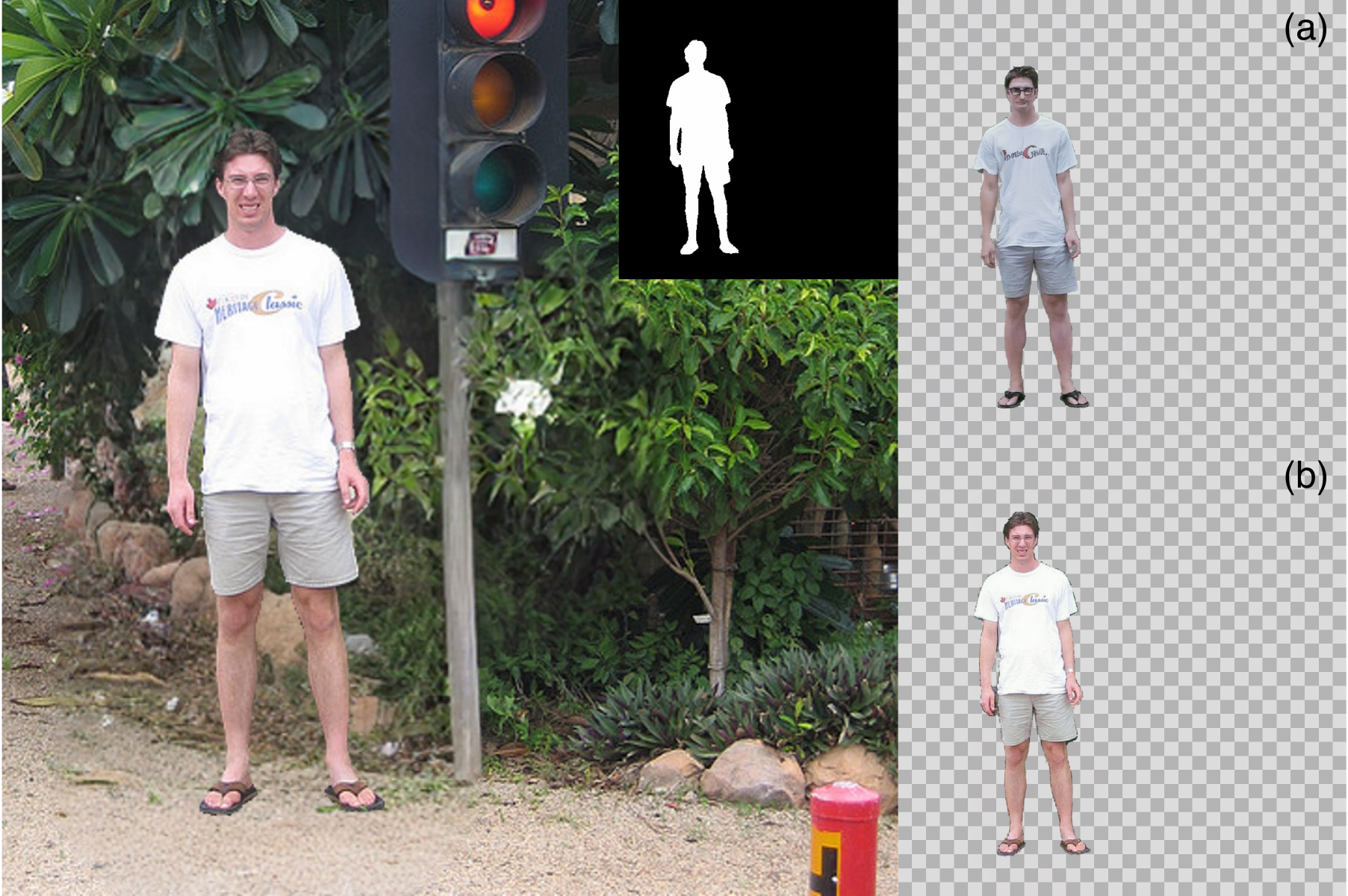}

    \vspace{-8pt}
    \caption{Comparison of Ablation Models Using Different Image-Mask Contexts for Image Layer Decomposition. We compare models trained with the standard inpainting-only context $c^{b}_{I-M}$  and our proposed dual-context design $\{c^{f}_{I-M}, c^{b}_{I-M}\}$. In (a), the foreground output is generated using only $c^{b}_{I-M}$, in (b), the foreground is generated using the combined context. The key difference is in content fidelity, model (a) tends to hallucinate or alter parts of the foreground, while model (b) better preserves the original object, demonstrating the importance of our dual-context design for faithful layer decomposition. }
    \label{fig:fgbg_mask}
    \vspace{-9pt}
\end{figure}

We also conduct a quantitative ablation study to evaluate the impact of our carefully designed image-mask context, as illustrated in Figure~\ref{fig:image-mask}. Specifically, we compare our full context design, using both foreground and background image-mask context $\{c^{f}_{I-M}, c^{b}_{I-M}\}$ , with the standard background-only image-mask context $c^{b}_{I-M}$ typically used in inpainting tasks. The results of this comparison are reported in Table~\ref{tab:abla}. When the foreground image-mask context $c^{f}_{I-M}$, is removed, we observe a drop of 0.26 dB in PSNR on the object removal task. More importantly, as shown in Figure~\ref{fig:fgbg_mask}, the model's ability to faithfully extract foreground objects degrades significantly, beginning to hallucinate or alter content within the extracted regions. These findings validate the motivation behind our combined foreground-background image-mask context design: the inclusion of $c^{f}_{I-M}$ plays a crucial role in mitigating hallucination and ensuring the faithful preservation of object regions during layer decomposition.

\begin{table}[t!]
    \centering
    \resizebox{1.0\linewidth}{!}{
    \begin{tabular}{@{}cccccc@{}} % Four columns
    \midrule
    \midrule
    & \makecell{\textbf{Method}} & PSNR $\uparrow$  & SSIM $\uparrow$  & LPIPS $\downarrow$  & FID $\downarrow$  \\
    \midrule
     % & CNI & xxx& xxx & xxx \\
     & $r_{128}$  & 26.34 & 0.92 & 0.09 & 33.92 \\
     & $r_{1024}$   & 27.15 & 0.92 & 0.09 & 27.32\\
     \midrule
     & -\textit{w/o} $c^{f}_{I-M}$ & 27.04 & 0.92 & 0.08 & 27.49 \\
     & -\textit{w/o} $c_{MM}$ & 27.16 & 0.93 & 0.08 & 28.02 \\
     \midrule
    & No synthetic foreground & 27.18 & 0.92 & 0.08 & 27.11 \\
     \midrule
     & Kontext & 26.22 & 0.92 & 0.09 & 36.14 \\
     \midrule
    & w/o edge & 27.19 & 0.92 & 0.08 & 27.61 \\
    & w/o seg & 27.22 & 0.93 & 0.08 & 28.16 \\
    & w/o depth & 27.19 & 0.93 & 0.08 & 28.10 \\
    \midrule
     & Ours & \textbf{27.30} & \textbf{0.93} & \textbf{0.08} & \textbf{25.97} \\
    \midrule
    \midrule
    \end{tabular}}

    \vspace{-9pt}

    \captionsetup{font=small}
    \caption{Ablation Study: Quantitative comparison on the object removal task to evaluate the effectiveness of our design, grouped by ablation type:  1) LoRA rank: $r_{128}$ and $r_{1024}$ denote LoRA ranks of 128 and 1024, respectively. 2) Context: -\textit{w/o} $c^{f}_{I-M}$: removes the foreground image-mask context, -\textit{w/o} $c_{MM}$: removes the multi-modal context. 3) The effect of removeing generated foregrounds from training \textit{No synthetic foreground}, 4) Base model: \textit{Kontext} replaces the inpainting backbone, 5) \textit{w/o edge}, \textit{w/o seg}, \textit{w/o depth} ablate each modality in the multi-modal context separately. }
    \label{tab:abla}

    \vspace{-20pt}

\end{table}

\subsubsection{The Impact of Multi-modal Context}
\begin{figure}[!t]
    \centering
    \includegraphics[width=.999\linewidth]{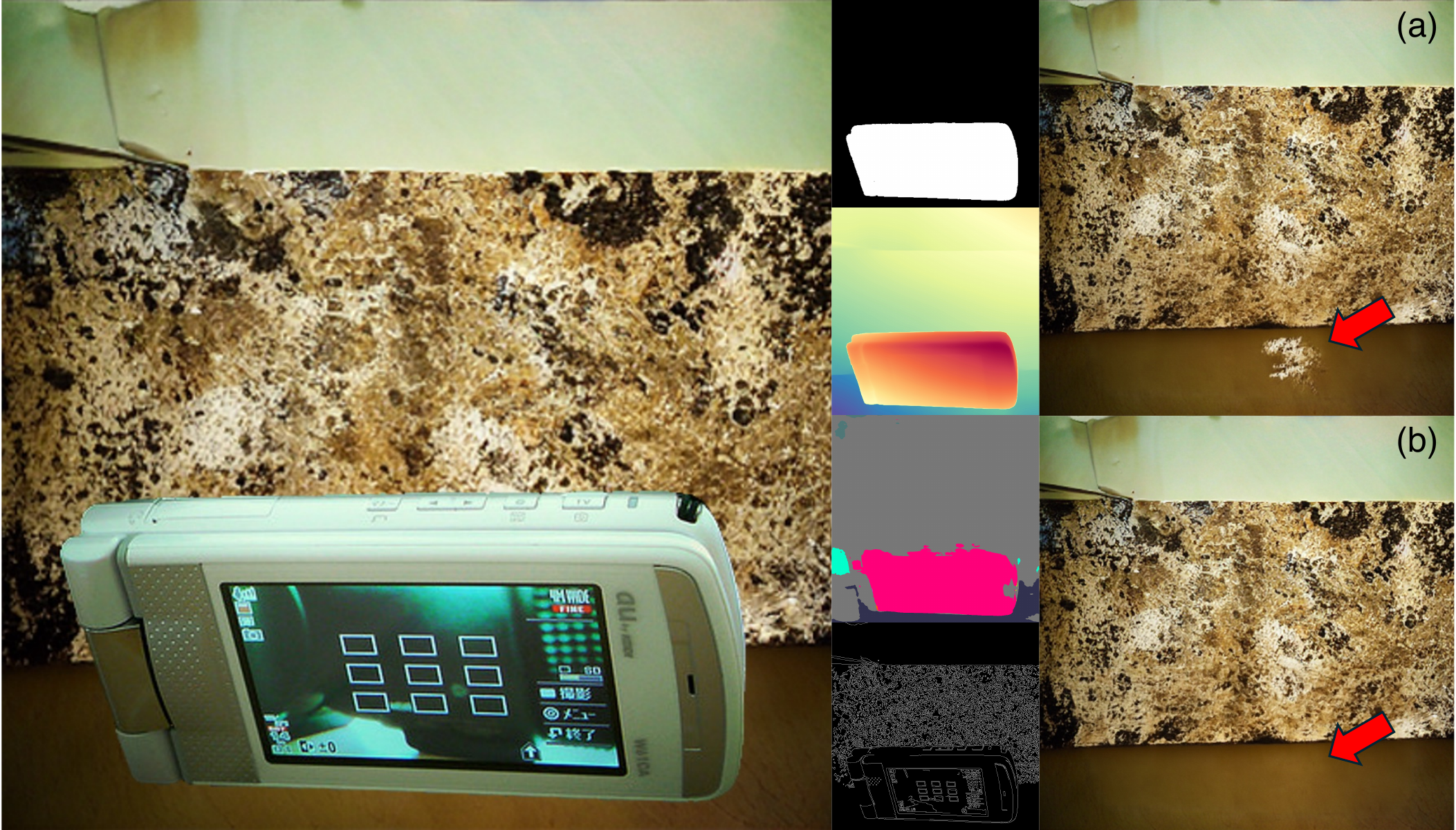}  

    \vspace{-8pt}
    \caption{Comparison of object removal with and without multi-modal context. We compare object removal under two settings: (a) without multi-modal context and (b) with multi-modal context. The left side shows the input image, mask, and multi-modal cues, including depth, segmentation, and edge maps; the right side shows the corresponding outputs. Incorporating multi-modal context clearly improves object removal by providing richer spatial and semantic cues, reducing hallucinations and producing more consistent, realistic reconstructions.}
    \label{fig:mm_abla}
    \vspace{-10pt}
\end{figure}

We present both quantitative and qualitative evaluations to assess the impact of multi-modal context. As shown in Table~\ref{tab:abla}, incorporating multi-modal context significantly improves quantitative performance on the object removal task, as evidenced by the drop in performance for -\textit{w/o} $c_{MM}$. We further ablate each modality individually to analyze its contribution to the final performance. More noticeably, Figure~\ref{fig:mm_abla} demonstrates that the inclusion of multi-modal cues, such as depth, segmentation, and edge maps, helps the model better understand the semantics of the fill-in regions. 

% This results in more accurate and visually consistent reconstructions. In contrast, without such context, the model is prone to hallucinations, often introducing unrealistic artifacts or structure in the removed regions. These results highlight the importance of multi-modal guidance in reducing hallucination and improving background realism after object removal.

\subsubsection{The Impact of LoRA Rank}
\label{sec:ablate-lora}
As a general trade-off in LoRA-based Parameter-Efficient Fine-Tuning (PEFT) frameworks, the choice of LoRA rank plays a crucial role. This trade-off is especially significant in our case due to two key factors: 1) Our base model is a pre-trained image inpainting diffusion model, originally trained for a different task. Adapting it to the novel task of image layer decomposition requires learning new capabilities while preserving its existing generative priors. 2) Our training data is entirely curated from public datasets and tools, which is inferior compared to the commercial-grade datasets used for training the original model.
Together, these factors motivate a careful investigation into how the LoRA rank impacts the adaptation process, seeking an optimal trade-off between preserving pre-trained priors and enabling the learning of new functionality. We report the quantitative ablation results in Table~\ref{tab:abla}. As shown: a smaller rank 128 is insufficient for the model to fully learn the new task of layer decomposition. A larger rank 1024 enables greater adaptation but overwrites pre-trained priors, leading to hallucinations and sub-optimal performance. Based on these findings, we select a LoRA rank of 256 for our final model configuration, achieving the best balance between adaptability and stability.
\section{Limitations and Future Work}
As shown in the Supplementary Material, our method fails on some complex images including cluttered objects, large occlusions, and objects held in fingers. We attribute it to the lack of such intricate samples in the synthetic training dataset, and believe it will be alleviated with better training data in the future work.
\section{Conclusion}
In this paper, we demonstrate that image layer decomposition shares intrinsic similarities with in/outpainting. Leveraging this insight, we show that unlike prior works, our approach can achieve generative layer decomposition using only publicly available data, a lightweight adaptation of an inpainting model, and carefully designed modules to preserve context and reduce hallucinations. Our method achieves state-of-the-art performance with significantly lower data and computational requirements.
{
    \small
    \bibliographystyle{ieeenat_fullname}
    \bibliography{main}
}

% WARNING: do not forget to delete the supplementary pages from your submission 
% \input{sec/X_suppl}

\end{document}